%% file: main.tex
\documentclass{article}

    \PassOptionsToPackage{numbers, compress}{natbib}

     \usepackage[preprint]{neurips_2024}

\usepackage[utf8]{inputenc} 
\usepackage[T1]{fontenc}    
\usepackage{hyperref}       
\usepackage{url}            
\hypersetup{colorlinks,linkcolor={blue},citecolor={blue},urlcolor={blue}}
\usepackage{booktabs}       
\usepackage{amsfonts}       
\usepackage{nicefrac}       
\usepackage{microtype}      
\usepackage{xcolor}         

\usepackage{booktabs}       
\usepackage{bm}
\usepackage{amsmath}
\usepackage{xcolor} 
\usepackage{algorithm,algorithmic}
\usepackage{graphicx}
\usepackage{caption}
\usepackage{subcaption}
\graphicspath{{images/}}
\usepackage{multirow}
\usepackage{longtable}
\usepackage{dirtree}
\usepackage{pdfpages}
\usepackage{amssymb}
\usepackage{enumitem}
\usepackage{wrapfig}
\newcommand{\algorithmicdefine}{\textbf{Define:}}
\newcommand{\DEFINE}[1]{\item[\algorithmicdefine] #1}

\renewcommand{\Re}[0]{\mathbb{R}}

\title{Second-Order Forward-Mode Automatic Differentiation for Optimization}

\author{%
  Adam D. Cobb\footnotemark[1]\thanks{Correspondence to \texttt{adam.cobb@sri.com}.},\quad 
  Atılım Güneş Baydin\footnotemark[2],\quad 
  Barak A. Pearlmutter\footnotemark[3],\quad 
    Susmit Jha\footnotemark[1] 
\\
\\
\footnotemark[1]\enspace Computer Science Laboratory, SRI International
\\ 
 \footnotemark[2]\enspace Department of Computer Science, University of Oxford\\
 \footnotemark[3]\enspace Department of Computer Science, National University of Ireland Maynooth
}

\begin{document}

\maketitle

\begin{abstract}
  This paper introduces a second-order hyperplane search, a novel optimization step that generalizes a second-order line search from a line to a $k$-dimensional hyperplane.
  This, combined with the forward-mode stochastic gradient method,
  yields a second-order optimization algorithm that consists of forward passes only, completely avoiding the storage overhead of backpropagation.
  Unlike recent work that relies on directional derivatives (or Jacobian--Vector Products, JVPs), we use hyper-dual numbers to jointly evaluate both directional derivatives and their second-order quadratic terms. As a result, we introduce forward-mode weight perturbation with Hessian information (FoMoH). We then use FoMoH to develop a novel generalization of line search by extending it to a hyperplane search. 
  We illustrate the utility of this extension and how it might be used to overcome some of the recent challenges of optimizing machine learning models without backpropagation. Our code is open-sourced at \url{https://github.com/SRI-CSL/fomoh}. 
\end{abstract}

\input{1_intro}

\input{6_related_work}

\input{2_preliminaries}

\input{3_fomoh}

\input{4_experiments}

\input{5_conclusion}

\begin{ack}
This material is based upon work supported by the United
States Air Force and DARPA under Contract No. FA8750-23-C-0519. Any opinions, findings
and conclusions or recommendations expressed in this material are those of the author(s) and do not necessarily reflect
the views of the United States Air Force and DARPA.
\end{ack}

\bibliographystyle{plainnat}
\bibliography{references}


\appendix

\input{7_appendix}


\end{document}

%% file: 1_intro.tex
\section{Introduction}

There is a growing interest in investigating the practical plausibility of forward-mode automatic differentiation (AD) as an alternative to 
reverse-mode AD (aka backpropagation) for neural network optimization.
Forward gradient descent (FGD) \citep{baydin2022gradients} relies on sampling tangent vectors to update function parameters. These parameters are updated by subtracting the tangents that are scaled by their directional derivatives. As a result, their approach only requires forward passes, and avoids the computation and memory costs associated with implementing the backwards pass of reverse-mode AD. While there is still an accuracy gap between forward-mode and reverse-mode optimization approaches \cite{silver2021learning}, there are several recent efforts that have explored alternative ways of leveraging forward-mode AD, with a focus of reducing the variance of the gradient estimator with the increase in dimensions \cite{ren2022scaling, fournier2023can}. 
Thus, the large performance gap between the forward-mode approaches and backpropagation (BP) has been shrinking.

An interesting direction to explore, which builds on the existing work in forward-mode AD for optimization, is the introduction of Hessian information.
Second-order derivative information provides optimization routines with information about the local curvature. These methods often require access to the Hessian, whose storage is not feasible for high-dimensional problems.
As an example, a function, $f: \Re^D\rightarrow\Re$, has a quadratic storage cost of $\mathcal{O}(D^2)$ and a linear compute cost of $\mathcal{O}(D)$ reverse-mode gradient evaluations to build the Hessian \cite{pearlmutter1994fast, dagreou2024howtocompute}. 
Instead of building the full Hessian through the use of reverse-mode AD, we introduce an approach incorporating Hessian information, known as FoMoH, which significantly outperforms the first-order gradient descent method, FGD. Furthermore, we demonstrate how our approach effectively bridges the gap between a line search and a full Newton step. By incorporating Hessian information, FoMoH leverages second-order derivatives to provide more accurate curvature approximations of the objective function. This allows the method to adaptively adjust the step size and direction with greater precision compared to a simple line search, which only uses gradient information. Simultaneously, FoMoH avoids the computational complexity of a full Newton step, which requires calculation of $(\nabla^2 f)^{-1}\mathbf{v}$ where $\nabla^2 f$ is the Hessian. As a result, our approach balances the efficiency of a line search with the accuracy of a full Newton method, offering a robust and versatile optimization technique.

The central contributions of this paper are as follows:
{
\setlength{\topsep}{0pt} 
\setlength{\partopsep}{0pt} 
\begin{itemize}[leftmargin=*]
    \item We introduce three new optimization approaches that use forward-mode second-order information.
    \item We show how one of these approaches, Forward-Mode Hyperplane Search, generalizes from a line search all the way to Newton's method, without any need for backpropagation.
    \item We demonstrate the performance of the proposed new approaches on optimization problems of different parameter sizes and difficulty to show the advantage of second-order information.
    \item We release an AD backend in PyTorch that implements nested forward AD and interfaces with PyTorch models: \url{https://github.com/SRI-CSL/fomoh}.
\end{itemize}
}

The rest of the paper is organized as follows. §\ref{sec:relwork} and §\ref{sec:prelim} summarize the related work and relevant preliminary information regarding AD, which is then followed by §\ref{sec:2AD} that outlines what is required to extend forward-mode AD to higher order derivatives. §\ref{sec:method} introduces FoMoH and the generalization to our forward-mode hyperplane search. Finally, §\ref{sec:exp} provides experimental results that explore the behavior of FoMoH. We then conclude in §\ref{sec:conc}.

%% file: 6_related_work.tex
\section{Related Work}\label{sec:relwork}

There has been considerable interest in developing approaches that avoid reverse-mode AD and its backwards pass. In moving away from backpropagation, it might be possible to build optimization algorithms that more closely align with biological systems \cite{bengio2015towards, hinton2022forward}, or enable neural networks to run on 
emerging hardware architectures, such as analog optical systems \cite{pierangeli2019large}. \citet{baydin2022gradients} introduced FGD as a possible replacement for backpropagation in neural network training by relying on weight perturbations. This removed the truncation error of previous weight perturbation approaches \cite{pearlmutter1994fast} by moving to forward-mode AD. While FGD is a promising backpropagation-free approach, 
the scaling of FGD to high-dimensional models is challenging due the variance of the gradient estimator.
This challenge has led to multiple efforts that have focused on reducing this variance \cite{ren2022scaling, silver2021learning, fournier2023can}. In particular, a common approach has  been to rely on local backpropagation steps to reduce the variance and/or to provide a better guess for the perturbation direction. In our work, we focus on second-order forward-mode optimization approaches, but highlight that these other approaches on variance reduction are orthogonal to our approach and could also be combined together and generalized to FoMoH.

\citet{becker1988improving} were one of the first to the use the second-order information to optimize neural networks. This approach leverages a local backpropagation step \cite{le1987modeles} to capture the diagonal Hessian terms resulting in what they call a ``pseudo-Newton step''. Both \citet[§6--§9]{lecun2002efficient} and \citet[§8.6]{Goodfellow-et-al-2016} provide discussions of the use of Hessian information for neural network training and cover Newton's method, as well as the Levenberg--Marquardt algorithm \cite{levenberg1944method, marquardt1963algorithm}, conjugate gradients, and BFGS. A key objective of these approaches is to investigate whether second-order information can be leveraged for neural network training without the prohibitively high cost of a full Hessian evaluation. Additionally, the research question then explored is whether the approximated Hessian is still good enough to give the desired advantage over first-order methods. Examples of effective approaches often rely on diagonal preconditioners, with some approximating or directly using the diagonal of the Hessian \cite{yao2021adahessian, le1987modeles, becker1988improving}, and others leveraging momentum and a variant of the diagonal of the empirical Fisher \cite{duchi2011adaptive, tieleman2012rmsprop, kingma2014adam}. While extending a gradient preconditioner to the full inverse Hessian is generally infeasible, there are approaches that use: block-diagonal approximations, low-rank approximations, and Krylov-subspace based approximations  \cite{le2010fast, vinyals2012krylov}. Additional references for these preconditioners can be found in \citet{martens2020new}. Finally, although it is challenging to scale second-order approaches to large neural network models, a recent work, Sophia \cite{liu2023sophia}, has managed to show success in using such an approach. Like previous works, they use a diagonal approximation of the Hessian. Importantly, they show that Sophia requires half the number of update steps as Adam \cite{kingma2014adam} to train a language model (GPT-2 \cite{radford2019language}). It is worth noting that none of the described second-order optimization methods rely solely on forward-mode AD like the one being proposed in this paper.

%% file: 2_preliminaries.tex
\section{Automatic Differentiation}\label{sec:prelim}
In this section we summarize the common definitions of forward-mode and reverse-mode automatic differentiation. For a complete introduction, please refer to \citet{griewank2008evaluating}. We use  $\mathbf{x}\in\Re^D$ to denote a column vector.

\textbf{Forward-Mode AD} \cite{wengert1964simple} applies the chain rule in the \textit{forward} direction. The forward-mode evaluation, $F(\bm{\theta}, \mathbf{v})$, requires an additional tangent vector, $\mathbf{v}\in\Re^D$, along with the parameter vector $\bm{\theta}\in\Re^D$ for a function $\boldsymbol{f}: \Re^D\rightarrow\Re^O$. The result of the evaluation, $[\boldsymbol{f}(\bm{\theta}), \nabla \boldsymbol{f}(\bm{\theta}) \mathbf{v}]$, is a function evaluation and the corresponding Jacobian vector product (JVP), where $\nabla \boldsymbol{f}(\bm{\theta})\in \Re^{O\times D}$. For a unidimensional output function, the JVP is the directional derivative,  $\nabla \boldsymbol{f}(\bm{\theta}) \cdot \mathbf{v}$. The time and space (memory cost) complexity 
are linear, both approximately twice that of a single function evaluation.\footnote{More precisely, the basic time complexity 
of a forward evaluation is constant $\in [2,2.5]$ times that of the function call \cite{griewank2008evaluating}.} A common implementation of forward-mode AD is to use dual numbers. A dual number $a + b \epsilon \in \mathbb{D}(\Re)$ contains a real (primal) component, $a\in\Re$, and a dual component, $b\in\Re$. We can think of this as representing a truncated Taylor series, $a+b\epsilon+\mathcal{O}(\epsilon^2)$, notationally simplified by the rule $\epsilon^2 = 0$. Using this, $f(a + b \epsilon) = f(a) + \nabla f(a) b \epsilon$. A simple example can be shown for the function, $f(a_1, a_2) = a_1 \times a_2$. Using dual numbers, $(a_1 + b_1 \epsilon)(a_2 + b_2 \epsilon)$, we retrieve the function evaluation and the corresponding well-known product rule: $a_1 a_2 + (a_1 b_2 + b_1 a_2) \epsilon$. This can be extended to multiple dimensions, and is the basis of forward-mode AD: lift all real numbers $\Re$ to dual numbers $\mathbb{D}(\Re)$.

\textbf{Reverse-Mode AD} requires both a forward pass and a reverse pass. The reverse-mode evaluation $R(\bm{\theta}, \boldsymbol{u})$, also requires the additional adjoint vector, $\boldsymbol{u}\in\Re^O$, which is often set to $1$ for scalar-valued functions. Using the same notation as for forward-mode, an evaluation of reverse mode results in the vector-Jacobian product, $\boldsymbol{u}^{\top} \nabla \boldsymbol{f}(\bm{\theta})$, as well as the function evaluation. When $\boldsymbol{u} = 1$, this results in the gradient $\nabla \boldsymbol{f}(\bm{\theta})$. Reverse-mode is required to store intermediate values on the forward pass that are used during the reverse pass. This results in a higher time and space complexity, that is higher computational cost and memory footprint per call of $R(\bm{\theta}, \boldsymbol{u})$. However, for the scalar-valued functions ($O=1$) that are common for ML optimization problems, only a single call of $R(\bm{\theta}, \boldsymbol{u})$ is needed to collect all the gradients compared to $D$ (dimension of inputs) calls of $F(\bm{\theta}, \mathbf{v})$. This is one of the key reasons for the widespread adoption of the reverse-mode AD in current gradient-based machine learning methods, despite the higher memory footprint.  

\section{Higher-Order Forward Mode Automatic Differentiation}\label{sec:2AD}
As described in the previous section, forward-mode automatic differentiation implementations can 
use dual numbers, $\bm{\theta} + \mathbf{v}\epsilon$. Dual numbers can be extended to truncate at a higher order, or to not truncate at all; and to allow nesting by supporting multiple distinct formal $\epsilon$ variables \cite{pearlmutter2007lazy}. Specifically focusing on second-order terms is often referred to as hyper-dual numbers (for example in the Aeronautics and Astronautics community \cite{fike2011development}). A hyper-dual number is made up from four components, which is written as $\bm{\theta} + \mathbf{v}_1\epsilon_1 + \mathbf{v}_2\epsilon_2 + \mathbf{v}_{12}\epsilon_1\epsilon_2$. In the same manner that we look at imaginary and real parts of complex numbers, we can look at the first derivative parts of a hyper-dual number by inspecting the $\epsilon_1$ and $\epsilon_2$ components, and we can look at the second derivative by inspecting the $\epsilon_1\epsilon_2$ component. To understand how this formulation arises, we can introduce the definitions $\epsilon_1^2 = \epsilon_2^2 = (\epsilon_1\epsilon_2)^2 = 0$ and replace the Taylor series expansion of a function, $f : \Re^D \rightarrow \Re$ around $\bm{\theta}$ with perturbation, $\mathbf{d}\in\Re^D$, with an evaluation of a hyper-dual number:\footnote{Note, we have left our function definition as being a scalar output for pedagogical reasons but nothing precludes a vector or matrix output, which is required for the composition of functions in most ML architectures.}
\begin{align*}
    f(\bm{\theta} + \mathbf{d}) &= f(\bm{\theta}) + \nabla f(\bm{\theta}) \mathbf{d} + \frac{1}{2} \mathbf{d}^{\top} \nabla^2 f(\bm{\theta})\mathbf{d} + \cdots \\
    f(\bm{\theta} + \mathbf{v}_1\epsilon_1 + \mathbf{v}_2\epsilon_2 + \mathbf{v}_{12}\epsilon_1\epsilon_2) &= f(\bm{\theta}) + \nabla f(\bm{\theta}) \mathbf{v}_1 \epsilon_1 + \nabla f(\bm{\theta}) \mathbf{v}_2 \epsilon_2 \\ & \quad + \nabla f(\bm{\theta}) \mathbf{v}_{12} \epsilon_1\epsilon_2 +  \mathbf{v}_1^{\top} \nabla^2 f(\bm{\theta})\mathbf{v}_2 \epsilon_1\epsilon_2+ \cdots.
\end{align*}
An alternative but isomorphic view is to regard $a+b\epsilon_1+c\epsilon_2+d\epsilon_1\epsilon_2 = (a+b\epsilon_1)+(c+d\epsilon_1)\epsilon_2$ as an element of $\mathbb{D}(\mathbb{D}(\Re))$, with subscripts to distinguish the inner vs outer $\mathbb{D}$s; from an implementation perspective, hyperduals can be regarded as inlining the nested structures into a single flat structure.

\subsection{Implications of Hyper-Dual Numbers for Machine Learning}

A function evaluation with hyper-dual numbers takes an input vector, $\bm{\theta} + \mathbf{v}_1\epsilon_1 + \mathbf{v}_2\epsilon_2 + \mathbf{0}\epsilon_1\epsilon_2$, with the $\epsilon_1\epsilon_2$ part set to zero. A typical setting to get exact gradient and Hessian elements is to set $\mathbf{v}_1 = \mathbf{e}_i$ and $\mathbf{v}_2 = \mathbf{e}_j$, where $\mathbf{e}_i$ and $\mathbf{e}_j$ are each a basis of one-hot unit vectors. Therefore, these basis vectors select the corresponding elements of the gradient and Hessian:  $$f(\bm{\theta}) + \nabla f(\bm{\theta})_i \epsilon_1 + \nabla f(\bm{\theta})_j \epsilon_2 + \nabla^2 f(\bm{\theta})_{ij} \epsilon_1\epsilon_2= f(\bm{\theta} + \mathbf{e}_i\epsilon_1 + \mathbf{e}_j\epsilon_2 + \mathbf{0}\epsilon_1\epsilon_2).$$
A single loop over the input dimension provides the exact gradient, whereas a nested loop provides the full Hessian. As a side note, a single loop also can give the Hessian vector product. This is done by setting one of the tangent vectors to the chosen vector, and looping through the basis for the other tangent vector.\footnote{While interesting, these results might not immediately seem attractive to the ML community. The Hessian calculation of a model with parameters, $\bm{\theta}\in\Re^D$, would require $D \cdot (D+1)/2$ function evaluations, whereas Hessian vector products are widely available in many AD libraries leveraging tricks such as a forward over reverse routine \cite{paszke2019pytorch, jax2018github}.} However, the key advantage is that with a single forward pass, we get both first order and second order information in the form of directional derivatives and the local curvature information respectively. We have already seen from \citet{baydin2022gradients} and follow up works \cite{ren2022scaling, fournier2023can} that despite the scalar form of the directional derivative, it can still be used to build optimization routines that appear competitive with backpropagation. 
In this paper, we investigate whether the additional access to local curvature through forward-mode AD can enable improvements over the current FGD algorithm.

\paragraph{Local Curvature: $\mathbf{v}_1^{\top} \nabla^2 f(\bm{\theta}) \mathbf{v}_2$.} 
The Hessian contains the curvature information at a point, $\bm{\theta}$, in the form of the second order partial derivatives. When we evaluate a function over hyper-dual numbers we get the bilinear form, $\mathbf{v}_1^{\top} \nabla^2 f(\bm{\theta}) \mathbf{v}_2$. This is a function that operates over two vectors, such that it is linear in each vector separately. The bilinear form 
is common in optimization routines, such as for conjugate gradient descent to assess whether two vectors are conjugate with respect to the Hessian. The value of $\mathbf{v}_1^{\top} \nabla^2 f(\bm{\theta}) \mathbf{v}_2$ tells us about how curvature covaries along the two vectors. In the case where $\mathbf{v}_1 = \mathbf{v}_2 = \mathbf{v}$, we arrive at the quadratic form that describes the curvature, which indicates the rate of change of the slope as you move in the direction of $\mathbf{v}$. The quadratic form also provides information on the convexity of the function at that point. If $\mathbf{v}^{\top} \nabla^2 f(\bm{\theta}) \mathbf{v} > 0,  \forall \mathbf{v} \in \Re^{D}_{>0}$, then the function is convex at that point and the Hessian is positive definite. The curvature also indicates the sensitivity of moving in certain directions. For example, when using gradients to optimize a function, taking a gradient step in a region of low curvature is likely to increase the value of the function. However the same sized step in a region of large curvature could significantly change the value of the function (for better or worse).

\paragraph{Computational Cost.} The cost of a single forward pass with hyper-dual numbers is of the same order of time and space complexity 
as the original function call. This reduces the memory cost compared to reverse-mode AD. The forward pass with hyper-dual numbers scales in the same way as the original function call to the number of parameters. However, there is a constant overhead price (no change in computational complexity) to be paid in the form of evaluating the second order terms throughout a forward pass. An example is that the addition of two scalars now requires $4$ additions, and multiplication now requires $9$ products and $5$ additions. However, unlike for reverse-mode, these intermediate values can be overwritten once they have been used to propagate gradient information.

%% file: 3_fomoh.tex
\section{Forward-Mode Optimization with Second Order Information}\label{sec:method}
We now introduce the three new optimization routines that incorporate forward-mode second order information:
{
\setlength{\topsep}{0em} 
\setlength{\partopsep}{0pt} 
\begin{enumerate}\itemsep0em
    \item Forward-Mode Line Search (FoMoH)
    \item Forward-Mode Line Search with Backpropagation (FoMoH-BP)
    \item Forward-Mode Hyperplane Search (FoMoH-$K$D)
\end{enumerate}
}

\subsection{Forward-Mode Line Search: FoMoH}
We introduce a new gradient-based optimization routine leveraging Forward-Mode automatic differentiation with Hessian information (FoMoH). We begin with the one dimensional case, which uses a second-order line search \citep{pearlmutter1994fast, Schraudolph-Graepel-2002}. For a given update direction $\mathbf{v}\sim \mathcal{N}(\mathbf{0}, \mathbf{I})$ and learning rate $\eta$, FoMoH normalizes all gradient steps by the curvature, giving:
\begin{equation}\label{eq:fomoh}
    \bm{\theta}' = \bm{\theta} + \eta \frac{\nabla f(\bm{\theta}) \cdot \mathbf{v}}{|\mathbf{v}^{\top} \nabla^2 f(\bm{\theta}) \mathbf{v}|} \mathbf{v}.
\end{equation}
The directional derivative in the numerator and the curvature (the quadratic form) in the denominator are both provided with \textbf{one forward-pass} of the  function $f(\bm{\theta} + \mathbf{v}\epsilon_1 + \mathbf{v}\epsilon_2 + \mathbf{0}\epsilon_1\epsilon_2)$. The normalization via this quadratic form results in accounting for the unit distance at the location $\mathbf{x}$ in the direction $\mathbf{v}$. Therefore, this update step takes into account the distance metric defined at $\bm{\theta}$. In regions of high curvature the step size will be smaller, which is a desirable behavior. Like Newton's method, setting $\eta\approx1.0$ seems to work in many cases, but by analogy with trust regions we suggest the inclusion of a learning rate, although we found the method to be reasonably robust to its value.

\subsection{Forward-Mode Line Search with Backpropagation: FoMoH-BP}

A line search starts with identifying a descent direction, followed by determining the value of the step size to move in that direction. Therefore, the second approach that we propose is to perform a line search on the gradient. Thus, this combines forward-mode and reverse-mode to build an optimization routine that provides the step-size for the ground truth gradient obtained from backpropagation. This additional step sets $\mathbf{v} = \nabla f(\bm{\theta})$ in Equation \eqref{eq:fomoh}. The result is an update step:
\begin{equation}\label{eq:fomoh-bp}
    \bm{\theta}' = \bm{\theta} + \eta \frac{\nabla f(\bm{\theta}) \cdot \nabla f(\bm{\theta})}{|\nabla f(\bm{\theta})^{\top} \nabla^2 f(\bm{\theta}) \nabla f(\bm{\theta})|} \nabla f(\bm{\theta}).
\end{equation}
Unlike FoMoH (and FoMoH-$K$D in the next section), FoMoH-BP includes a single reverse-mode AD evaluation. Specifically, this requires a single backpropagation step followed by forward-mode step that sets the tangent vectors to the gradient.

\subsection{Forward-Mode Hyperplane Search: FoMoH-$K$D}
The final algorithm that we propose is the Forward-Mode $K$-Dimensional Hyperplane Search, FoMoH-$K$D. Rather than performing a second order line search along direction $\mathbf{v}$, we perform a $K$-dimensional hyperplane search. Starting with the $K=2$ example, if we take two search directions, $\mathbf{v}_1$ and $\mathbf{v}_2$, we can build a $2\times 2$ matrix to form a Hessian in the plane defined by $\bm{\theta} + \kappa_1 \mathbf{v}_1 + \kappa_2 \mathbf{v}_2$. We evaluate a function, $f(\cdot)$, with a hyper-dual number using the pairs $\{\mathbf{v}_1,\mathbf{v}_1\}$, $\{\mathbf{v}_1,\mathbf{v}_2\}$, and $\{\mathbf{v}_2,\mathbf{v}_2\}$ for the $\epsilon_1, \epsilon_2$ coefficients. The result is the Hessian, $\mathbf{\tilde{H}}_{2\times2}$, in the $2\times2$ plane, and corresponding step sizes, $\kappa_1$ and $\kappa_2$, to take in each search direction:
\begin{align}
\mathbf{\tilde{H}}_{2\times2} &= \left[\begin{array}{cc}
   \mathbf{v}_1^{\top} \nabla^2 f(\bm{\theta}) \mathbf{v}_1  & \mathbf{v}_1^{\top} \nabla^2 f(\bm{\theta}) \mathbf{v}_2 \\
   \mathbf{v}_2^{\top} \nabla^2 f(\bm{\theta}) \mathbf{v}_1  & \mathbf{v}_2^{\top} \nabla^2 f(\bm{\theta}) \mathbf{v}_2
\end{array}\right],
&
\left[\begin{array}{c}
    \kappa_1  \\
      \kappa_2
\end{array}\right] &= \mathbf{\tilde{H}}_{2\times2}^{-1} \left[\begin{array}{c}
    \mathbf{v}_1^{\top} \nabla f(\bm{\theta})  \\
      \mathbf{v}_2^{\top} \nabla f(\bm{\theta})
\end{array}\right]
\end{align}
As a result, we formulate a new update step,
$$
\bm{\theta}' = \bm{\theta} + \kappa_1 \mathbf{v}_1 + \kappa_2 \mathbf{v}_2.
$$
We then extend the above result to any $K$-dimensional hyperplane by sampling $K$ search directions and evaluating the corresponding update step:
\begin{equation}\label{eq:generalFomoh}
    \bm{\theta}' = \bm{\theta} + \eta \sum_{k=1}^K \kappa_k \mathbf{v}_k.
\end{equation}
This resulting generalized hyperplane update step allows one to trade-off computational cost with the size of the search space. For example, the cost of evaluating a $K$-dimensional Hessian and then invert it is $\mathcal{O}(K^3)$, which is feasible for small enough $K$.\footnote{For instances where $\mathbf{\tilde{H}}_{K\times K}$ is not invertible, we add jitter to the diagonal. This seems to work well.} Our new forward-mode hyperplane search, FoMoH-$K$D, opens up the possibility of transitioning between a line search, when $K=1$, all the way to a full Newton step, when $K=D$, which we demonstrate in §\ref{sec:rosen}. The pseudo-code for a single update step is given in Algorithm \ref{alg:hyperplane_step}. The overall FoMoH-$K$D routine is given in Algorithm \ref{alg:FoMoH}.

\begin{algorithm}
\caption{FoMoH-$K$D, hyperplane update step for function, $f$, and parameters, $\bm{\theta} \in \Re^D$.}
\label{alg:hyperplane_step}
\begin{algorithmic}
\DEFINE $\mathrm{HyperPlaneStep}(f, \bm{\theta}, K)$
\STATE \textbf{Set:} $N = (K^2 + K)/2$, $\bm{\Theta} \in \Re^{N\times D}$, $\boldsymbol{V} \in \Re^{K\times D}$, $\boldsymbol{V}_1 \in \Re^{N\times D}$, $\boldsymbol{V}_2 \in \Re^{N\times D}$
\STATE \% For loop vectorized in code.
\FOR{$n = 1, \dots, N$} 
    \STATE $\bm{\Theta}[n, :] = \bm{\theta}$ \quad \quad \% Repeat current parameter values for vectorized evaluation.
\ENDFOR
\FOR{$k = 1, \dots, K$}
    \STATE $\boldsymbol{V}[k, :] = \mathbf{v}\sim \mathcal{N}(\mathbf{0}, \mathbf{I})$ \quad \quad \% Sample $K$ tangent vectors to build hyperplane.
\ENDFOR
\STATE \% Vectorize tangent vectors for evaluation of all $N$ elements of $\mathbf{\tilde{H}}_{K\times K}$.
\STATE $l = 1$
\FOR{$i = 1, \dots, K$} 
    \FOR{$j = i, \dots, K$} 
        \STATE $\boldsymbol{V}_1[l, :] = \boldsymbol{V}[i, :]$
        \STATE $\boldsymbol{V}_2[l, :] = \boldsymbol{V}[j, :]$
        \STATE $l=l+1$
    \ENDFOR
\ENDFOR
\STATE $\mathbf{z}_0 + \mathbf{z}_1\epsilon_1 + \mathbf{z}_2 \epsilon_2 + \mathbf{z}_{12}\epsilon_1\epsilon_2  = f(\bm{\Theta} + \boldsymbol{V}_1\epsilon_1 + \boldsymbol{V}_2 \epsilon_2 + \mathbf{0}\epsilon_1\epsilon_2)$
\STATE \% Build $\mathbf{\tilde{H}}_{K\times K}$ and directional derivatives vector, $\mathbf{\tilde{G}}_{K}$, in order to evaluate Eq. \eqref{eq:generalFomoh}.
\STATE \textbf{Set:} $\mathbf{\tilde{H}}_{K\times K} \in\Re^{K\times K}$, $\mathbf{\tilde{G}}_{K}\in\Re^{K\times 1}$
\STATE $l = 1$
\FOR{$i = 1, \dots, K$} 
    \STATE $\mathbf{\tilde{G}}_{K}[i] = \mathbf{z}_1[l]$
    \FOR{$j = i, \dots, K$} 
        \STATE $\mathbf{\tilde{H}}_{K\times K}[i,j] = \mathbf{z}_{12}[k]$
        \STATE $\mathbf{\tilde{H}}_{K\times K}[j,i] = \mathbf{z}_{12}[k]$ 
        \STATE $l=l+1$
    \ENDFOR
\ENDFOR
\STATE \% Returns update direction of vector size $D$
\RETURN $\sum_k( (- \mathbf{\tilde{H}}_{K\times K}^{-1} \mathbf{\tilde{G}}_{K})[k] \cdot \boldsymbol{V}[k,:])$ 
\end{algorithmic}
\end{algorithm}

%% file: 4_experiments.tex
\section{Experiments}\label{sec:exp} 

\subsection{Rosenbrock Function}\label{sec:rosen}

In this section, we test FoMoH and its variants on the Rosenbrock function \cite{rosenbrock1960automatic} which has a global minimum at $f(\mathbf{1}) = 0$. The Rosenbrock function, $f(\bm{\theta}) = \sum_{i=1}^{D-1} (100 (\theta_{i+1} - \theta_i^2)^2 + (1 - \theta_i)^2)$, is designed to be a challenging test case for non-convex optimization where the solution falls inside a narrow valley. The learning rate for the FoMoH variants is set to $1.0$ for the Rosenbrock experiments. 

\begin{figure}[h!]
    \centering
    \begin{subfigure}[t]{0.49\textwidth}
        \centering
        \includegraphics[width=\textwidth]{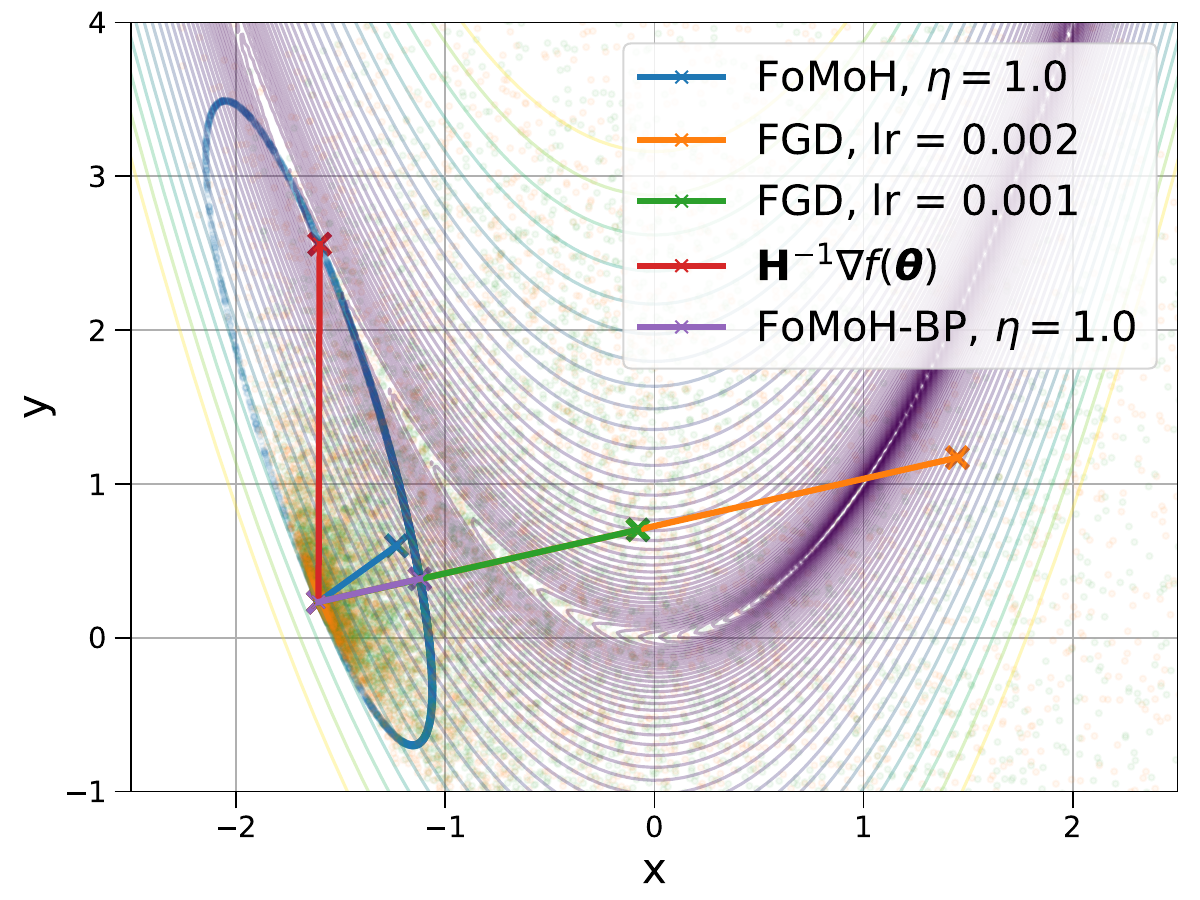}
        \caption{Expected step taken by the stochastic approaches of FoMoH and FGD. Included is a single Newton step for reference. The samples show how the curvature constrains the step size, compared to the sensitivity of FGD to the learning rate.}
        \label{fig:single_step}
    \end{subfigure}
    \hfill 
    \begin{subfigure}[t]{0.49\textwidth}
        \centering
        \includegraphics[width=\textwidth]{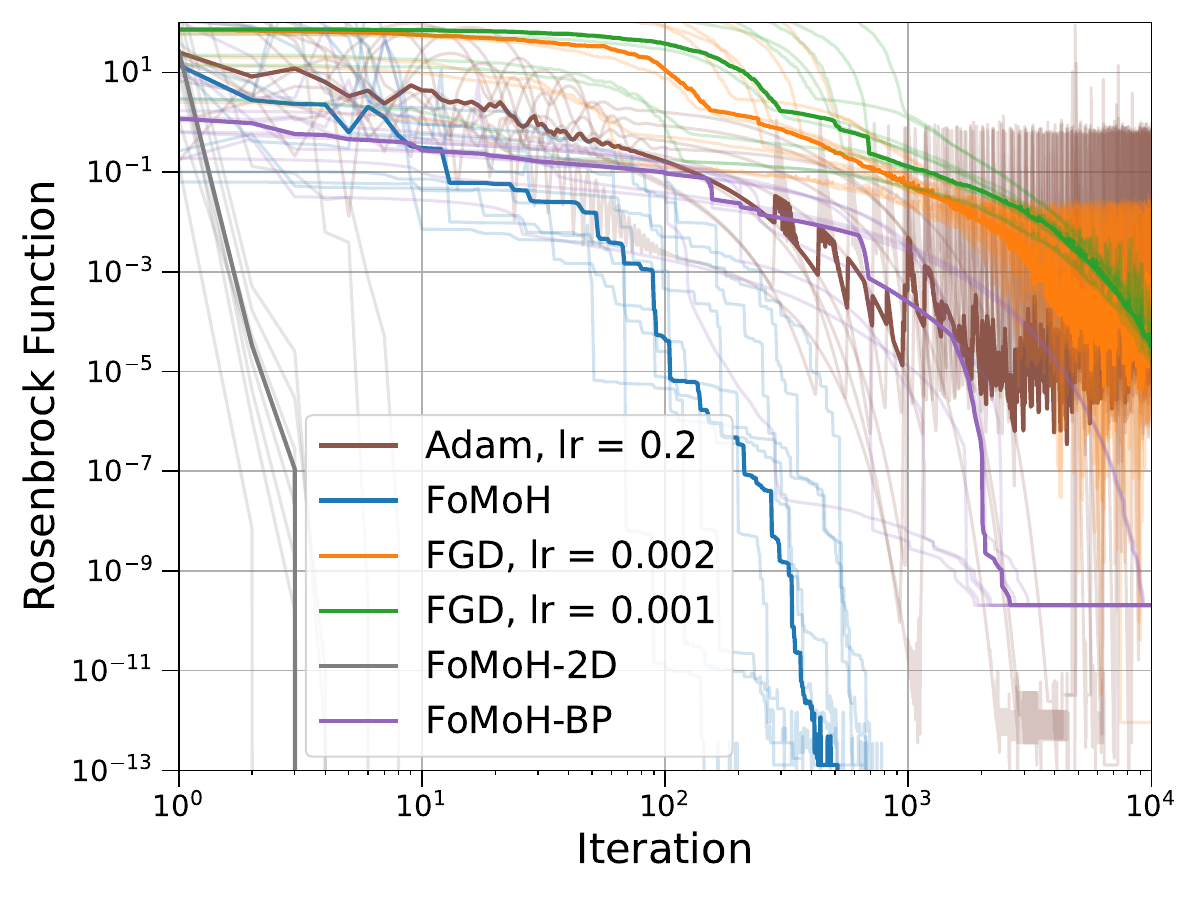}
        \caption{Comparison of the stochastic approaches in minimization of the Rosenbrock function. Average performance (median) is shown over $10$ random initial conditions. FoMoH outperforms all first-order approaches, with FoMoH-2D converging in orders of magnitude faster than all methods.}
        \label{fig:rosen_comp}
    \end{subfigure}
    \caption{Results over the 2D Rosenbrock function.}
    \label{fig:main}\vspace{-0.1in}
\end{figure}

As an initial illustration of the behavior of each forward-mode approach, we show how a single step looks from a randomly chosen starting point in Figure \ref{fig:single_step} for FGD, FoMoH, and FoMoH-BP. We plot the expected (average) step across $10{,}000$ samples for all approaches in the 2D Rosenbrock function. These steps are shown with solid lines. For each approach, we also plot the sampled steps by superimposing a scatter plot in the corresponding approach's color. All methods are compared to the Newton step, $(\nabla^2 f(\bm{\theta}))^{-1}\,\nabla f(\bm{\theta})$ shown in red. For FGD, we see that the expected descent direction is the same as the gradient at that point, hence the alignment with FoMoH-BP that directly calculates the gradient. This plot highlights the reliance on a well-chosen learning rate for FGD, whereas FoMoH-BP's step size is automatically normalized by the local curvature along the gradient. FoMoH (blue), on the overhand, has a descent direction that differs from the gradient and is governed by the distribution of samples that fall on the ellipse defined by the Hessian, $\nabla^2 f(\bm{\theta})$. For this point, the Newton step is the descent direction that falls on this ellipse and corresponds to the local minimum of the quadratic approximation. Another insight gained from this figure is that the variance of FGD's descent direction is less constrained than FoMoH's descent direction (blue), where the sample direction is controlled by the local Hessian. As a final note, we do not plot FoMoH-2D in Figure \ref{fig:single_step} as it directly aligns with the Newton step, with very little variance (see §\ref{sec:app}, Figure \ref{fig:rosenhist}).

In Figure \ref{fig:rosen_comp}, we now compare the performance of the competing optimization routines for the 2D Rosenbrock function. These results are shown for the same $10$ randomly sampled starting locations for all approaches initialized with different random seeds. We highlight with the thicker line the median performing run. Both axes are on the log scale and show the significant advantage of FoMoH-$K$D, with $K=2$. We also note the advantage (at least for this 2D example) of all FoMoH approaches that use second-order information. Additionally, the two forward-mode only approaches actually outperform the optimization routines that include backpropagation.

\begin{figure}[h!]
  \centering
  \includegraphics[width=0.5\textwidth]{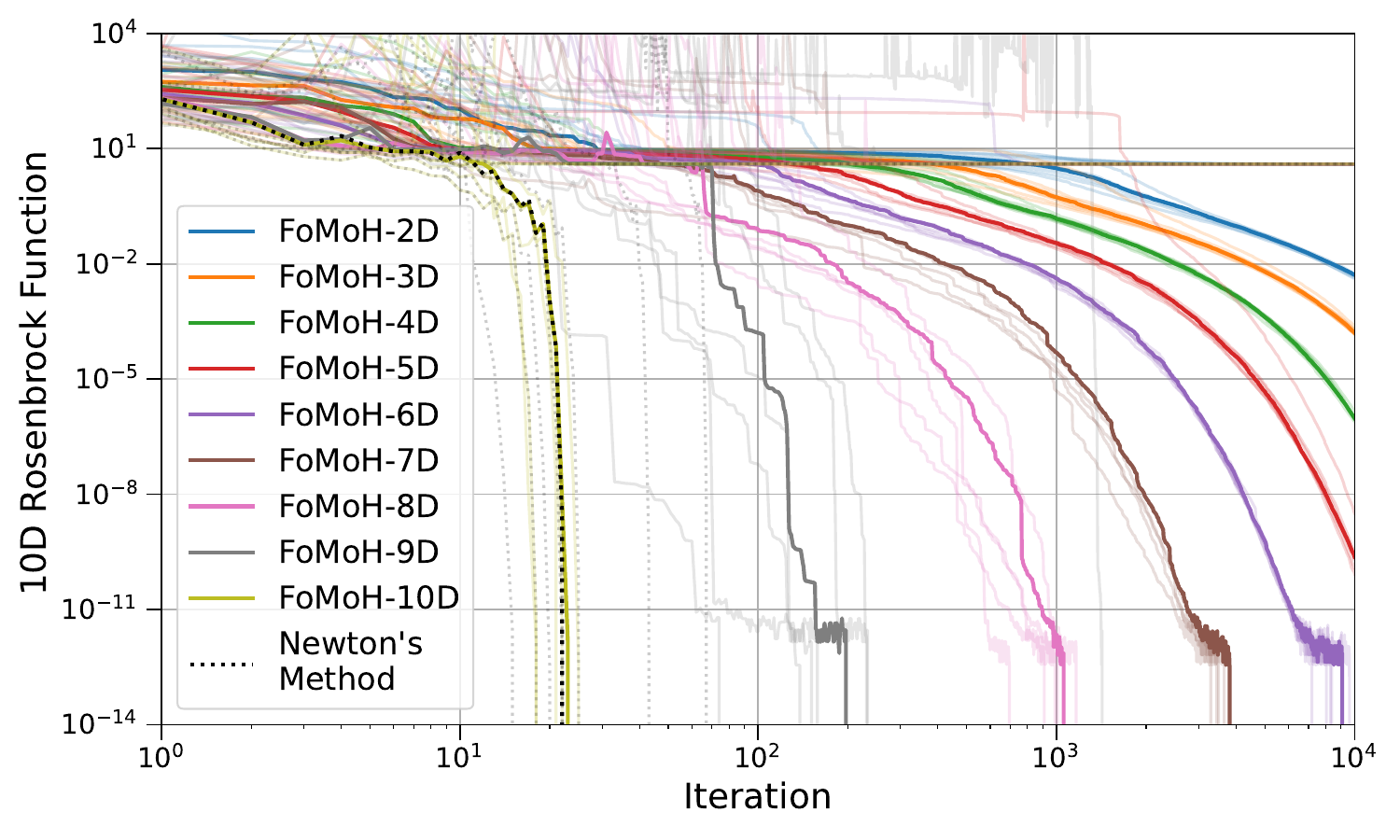}
  \caption{Performance of FoMoH-$K$D for $K=2\dots10$ on the 10D Rosenbrock function. Solid lines represent the median, with transparent lines corresponding to the each of the 10 random seeds. There is a clear pattern of higher dimensions performing better, with the performance of $K=10$ coinciding with Newton's Method (black dotted line).}
        \label{fig:rosendim}
\end{figure}

\paragraph{10D Rosenbrock Function.} We now focus on the performance of FoMoH-$K$D as we increase $K$ from $2$ to the input dimension of the function. Figure \ref{fig:rosendim} shows this comparison for the 10D Rosenbrock function, where we use 10 random initializations for the different $K$. The median performance for each $K$ is then shown, where we see a perfect ordering of performance that aligns with the dimension of the hyperplane. The best performing FoMoH-$K$D is for $K=D$, with the worst corresponding to the lowest dimension implemented, $K=2$. Overall this figure highlights how FoMoH-$K$D trends towards Newton's method as $K$ tends to $D$, where we actually see the median performances of FoMoH-10D and Newton's method aligned. 

\subsection{Logistic Regression}\label{sec:lr}
We now compare the performance of FoMoH for logistic regression applied to the MNIST dataset \cite{lecun1998gradient}. Table \ref{tab:LogReg_results} displays mean and standard deviation performance for each approach, where the forward-mode-only approaches are highlighted separately from the methods that include reverse-mode steps. Additional details on hyperparameter optimization are included in Appendix \ref{app:logreg}. Figure \ref{fig:FM-LR} displays the training and validation curves for both accuracy and negative log-likelihood (NLL) for the forward-mode approaches (Figure \ref{fig:LR} in appendix includes additional reverse-mode approaches). We see an improvement in speed of convergence as $K$ increases. However, we also see that FoMoH and FoMoH-$K$D, with a fixed learning rate, degrade in performance after reaching their local minimum (NLL) or maximum (accuracy). We therefore introduce a learning rate scheduler to improve on this behavior. For this task, FGD is competitive with the FoMoH variants but is slower to converge. In §\ref{sec:cnn} we show how FGD degrades in performance for a larger parameter space.

\begin{table}[!h]
\caption{Logistic regression results for MNIST. When comparing the forward-mode-only approaches in the upper section of the table, we see improvement in performance with increasing hyperplane dimension for FoMoH-$K$D. For this logistic regression example, FoMoH-3D and FoMoH-2D with learning rate schedulers, are competitive with FGD. However we will see this result change with a larger dimensional problem in §\ref{sec:cnn}. Both reverse-mode approaches in the lower section of the table have similar performance, and are included for reference.}
\label{tab:LogReg_results}
\begin{center}
\begin{scriptsize}
\begin{sc}
\begin{tabular}{lcccc}
\toprule
Approach & Training Loss & Validation Loss & Training Accuracy & Validation Accuracy \\
\midrule
FGD  & $0.2976 \pm 0.0007$ & $\mathbf{0.2949 \pm 0.0017}$ &
$0.9154 \pm 0.0003$ & $0.9163 \pm 0.0019$ \\
FoMoH  & $0.3223 \pm 0.0013$ & $0.3186 \pm 0.0019$ &
$0.9073 \pm 0.0011$ & $0.9110 \pm 0.0021$ \\
FoMoH (LR-Sch.) & $0.3192 \pm 0.0012$ & $0.3160 \pm 0.0025$ &
$0.9085 \pm 0.0011$ & $0.9118 \pm 0.0021$ \\
FoMoH-2D  & $0.3010 \pm 0.0018$ & $0.3015 \pm 0.0031$ &
$0.9144 \pm 0.0009$ & $0.9149 \pm 0.0015$ \\
FoMoH-2D (LR-Sch.) & $0.2921 \pm 0.0014$ & $0.2951 \pm 0.0027$ &
$0.9174 \pm 0.0005$ & $\mathbf{0.9170 \pm 0.0010}$ \\
FoMoH-3D & $0.3449 \pm 0.0017$ & $0.3343 \pm 0.0034$ &
$0.8999 \pm 0.0008$ & $0.9036 \pm 0.0023$ \\
FoMoH-3D (LR-Sch.) & $\mathbf{0.2893 \pm 0.0017}$ & $0.3054 \pm 0.0034$ &
$\mathbf{0.9195 \pm 0.0007}$ & $0.9153 \pm 0.0010$ \\
\midrule
FoMoH-BP & $0.2312 \pm 0.0001$ & $0.2679 \pm 0.0003$ &
$0.9366 \pm 0.0001$ & $0.9267 \pm 0.0002$ \\
Backpropagation  & $0.2265 \pm 0.0000$ & $0.2710 \pm 0.0001$ &
$0.9381 \pm 0.0001$ & $0.9267 \pm 0.0003$ \\
\bottomrule
\end{tabular}
\end{sc}
\end{scriptsize}
\end{center}
\vskip -0.2in
\end{table}

\begin{figure}[h!]
    \centering
        \centering
        \includegraphics[width=\textwidth]{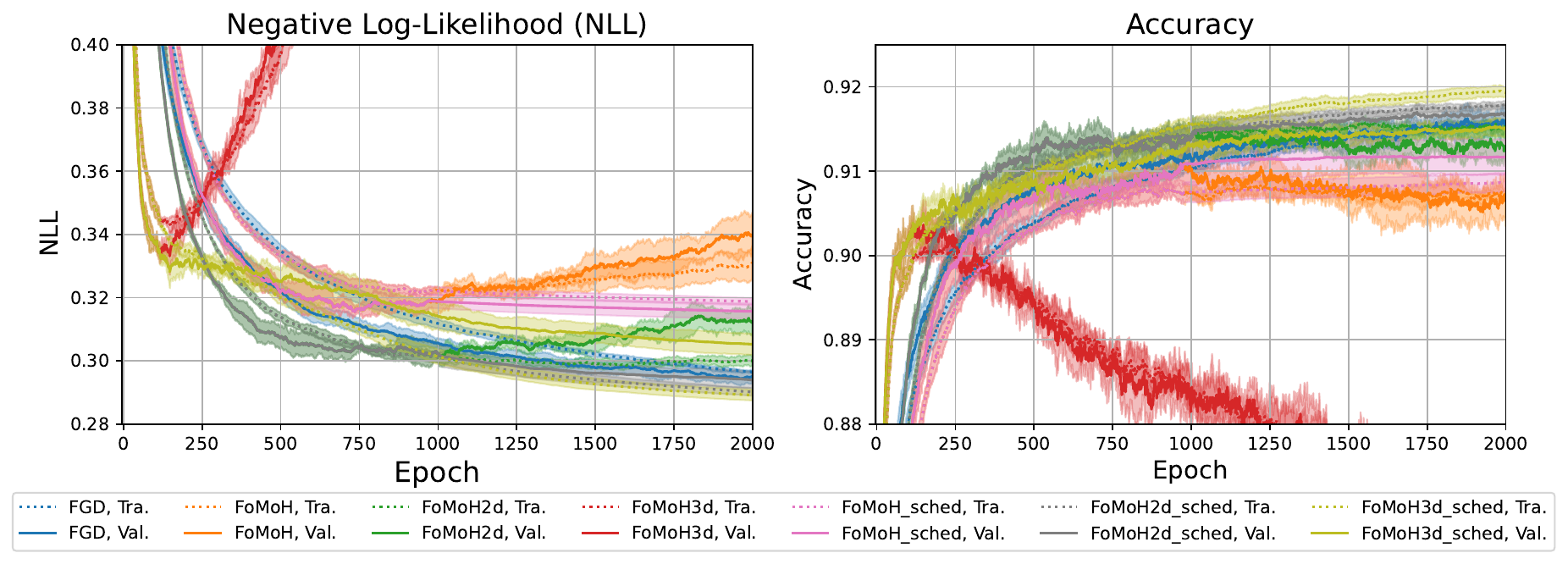}
        \caption{Forward-mode training and validation curves for the logistic regression model on the MNIST dataset. Average and standard deviation is shown for five random initializations.}
        \label{fig:FM-LR}\vspace{-0.1in}
\end{figure}

\subsection{Convolutional Neural Network}\label{sec:cnn}
\begin{figure}[h!]
    \centering
        \centering
        \includegraphics[width=\textwidth]{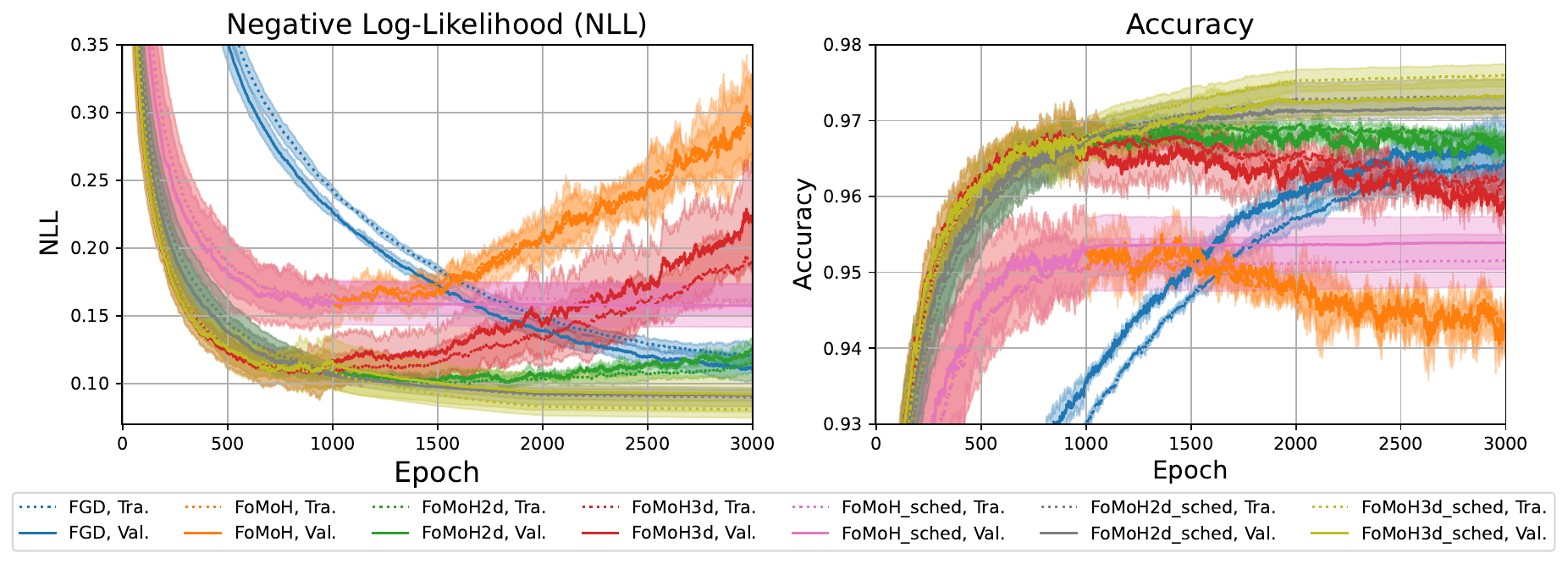}
        \caption{Forward-mode training and validation curves for the CNN on the MNIST dataset. Average and standard deviation is shown for three random initializations. Note how FGD (blue) is much slower to converge, with FoMoH-$K$D improving in performance with increasing $K$.}
        \label{fig:FM-CNN}
\end{figure}

We now move from a model with $7{,}850$ parameters to a convolutional neural network (CNN) with $431{,}080$ parameters. As before we use the MNIST dataset and leave the details on hyperparameter optimization using Bayesian optimization to the Appendix \ref{app:cnn}. 
Both Table \ref{tab:CNN_results} and Figure \ref{fig:FM-CNN} highlight the advantage of FoMoH-$K$D over FGD. For the larger parameter space FGD requires more epochs to converge compared to all FoMoH variants. The learning rate scheduler further improves FoMoH and FoMoH-$K$D by helping to avoid getting stuck in low performance regions. Here, we see the clear trend that the best performing forward-mode-only approach comes from the largest $K$, which was $K=3$ for this experiment. As expected, both optimizers FoMoH-BP and Backpropagation outperform the forward-mode-only approaches. Overall, these results highlight that second-order information helps scale the performance of forward-mode optimization to larger dimensions.

\begin{table}[!h]
\caption{CNN results for MNIST. The forward-mode-only approaches in the upper section of the table show that FoMoH's performance improves with the dimension of $K$, especially when used with the learning rate scheduler. FoMoH-3D outperforms FoMoH-2D, FoMoH, and FGD. The reverse-mode approaches in the lower section outperform forward-mode, with BP slightly better than FoMoH-BP.}
\label{tab:CNN_results}
\begin{center}
\begin{scriptsize}
\begin{sc}
\begin{tabular}{lcccc}
\toprule
Approach & Training Loss & Validation Loss & Training Accuracy & Validation Accuracy \\
\midrule
FGD  & $0.1211 \pm 0.0097$ & $0.1104 \pm 0.0086$ &
$0.9641 \pm 0.0038$ & $0.9677 \pm 0.0030$ \\
FoMoH & $0.1663 \pm 0.0069$ & $0.1571 \pm 0.0092$ &
$0.9518 \pm 0.0011$ & $0.9550 \pm 0.0002$ \\
FoMoH (LR-Sch.)  & $0.1617 \pm 0.0119$ & $0.1575 \pm 0.0158$ &
$0.9515 \pm 0.0035$ & $0.9539 \pm 0.0034$ \\
FoMoH-2D  & $0.1015 \pm 0.0041$ & $0.1016 \pm 0.0066$ &
$0.9691 \pm 0.0011$ & $0.9693 \pm 0.0020$ \\
FoMoH-2D (LR-Sch.)  & $0.0900 \pm 0.0070$ & $0.0913 \pm 0.0041$ &
$0.9731 \pm 0.0022$ & $0.9718 \pm 0.0014$ \\
FoMoH-3D & $0.1085 \pm 0.0105$ & $0.1073 \pm 0.0133$ &
$0.9674 \pm 0.0022$ & $0.9686 \pm 0.0028$ \\
FoMoH-3D (LR-Sch.)  & $\mathbf{0.0809 \pm 0.0061}$ & $\mathbf{0.0923 \pm 0.0132}$ &
$\mathbf{0.9759 \pm 0.0014}$ & $\mathbf{0.9734 \pm 0.0022}$ \\
\midrule
FoMoH-BP & $0.0093 \pm 0.0016$ & $0.0310 \pm 0.0006$ &
$0.9981 \pm 0.0005$ & $0.9903 \pm 0.0003$ \\
Backpropagation  &$0.0053 \pm 0.0034$ & $0.0329 \pm 0.0032$ &
$0.9990 \pm 0.0009$ & $0.9909 \pm 0.0004$ \\
\bottomrule
\end{tabular}
\end{sc}
\end{scriptsize}
\end{center}
\end{table}

%% file: 5_conclusion.tex
\section{Conclusion, Broader Impact, and Limitation Discussion}\label{sec:conc}
\label{sec:dis}

The results in §\ref{sec:exp} highlight the potential of the use of second-order forward-mode AD for optimization tasks. For the Rosenbrock function, we illustrated the behavior of our three new optimization routines: FoMoH, FoMoH-BP, and FoMoH-$K$D, and we compared them to Newton's method. In particular we were able to show that as one increases the hyperplane dimension of FoMoH-$K$D the method tends to Newton's method, without the need for any backpropagation. This significant result is shown in Figure \ref{fig:rosendim}. For the learning tasks of logistic regression and CNN classification, we see how the first-order optimization approach of FGD degrades with increasing dimension of the parameter space. We do not see this degradation for FoMoH-$K$D, and we also observe that the second-order information means fewer epochs are needed to reach a better performance. This  has the broader impact of improving efficiency, reducing cost, and increasing accuracy in ML optimization routines. 

In conclusion, we  introduced a novel approach that uses second-order forward-mode AD for optimization. We have introduced: forward-mode line search (FoMoH); forward-mode line search with Backpropagation (FoMoH-BP); and forward-mode hyperplane search (FoMoH-$K$D). We have shown how these approaches compare to the previous first-order forward-mode approach of FGD, as well stochastic gradient descent for multiple optimization problems over a wide range of dimensions. Furthermore, FoMoH-$K$D behaves closer to the performance of Newton's method as $K$ increases. In addition to contributing the new second-order forward-mode optimization routines, we provide a Python package that implements the AD backend and interfaces with PyTorch.
 Our work is a further step in the direction 
 of showing the value of cheap second-order information in optimization with the need to scale to even larger dimensions.
 We expect that future work will be able to mix the advantages gained from first-order approaches with that of second-order approaches.

%% file: 7_appendix.tex
\newpage

\section{Additional Results}\label{sec:app}

\subsection{FoMoH-$K$D Algorithm}
\begin{algorithm}
\caption{FoMoH-$K$D}
\label{alg:FoMoH}
\begin{algorithmic}
\REQUIRE Objective function $f(\bm{\theta})$, initial point $\bm{\theta}$, step size $\eta$, hyperplane dimension $K$.
\FOR{$t = 0, 1, 2, \dots$ until convergence}
    \STATE $\mathbf{d} = \mathrm{HyperPlaneStep}(f, \bm{\theta}, K)$ \% See Algorithm \ref{alg:hyperplane_step}
    \STATE $\bm{\theta} = \bm{\theta} + \eta\mathbf{d}$
\ENDFOR
\end{algorithmic}
\end{algorithm}

\subsection{Rosenbrock Function}
\begin{figure}[h!]
    \centering
        \centering
        \includegraphics[width=\textwidth]{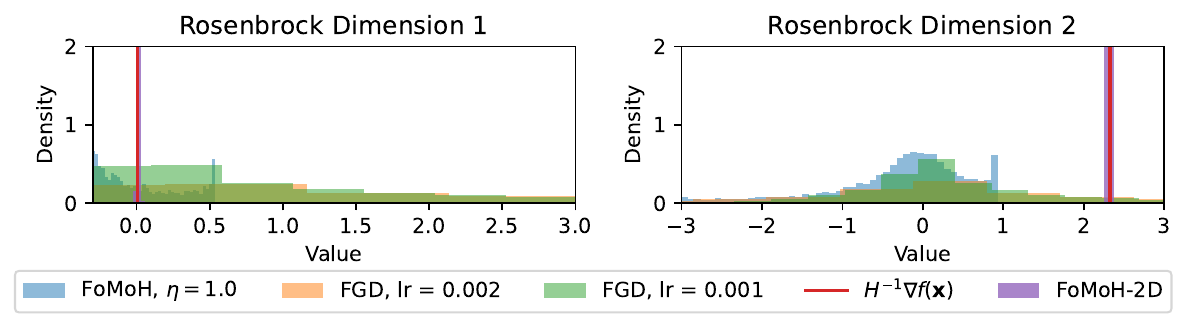}
        \caption{Histogram over expected step taken by the stochastic approaches of FoMoH, FGD, and FoMoH-2D corresponding to Figure \ref{fig:single_step}. Noteworthy is that the variance of the 2D hyperplane search step is significantly smaller and expectation is close to Newton step.}
        \label{fig:rosenhist}
\end{figure}

\subsection{Logistic Regression}\label{app:logreg}

Table \ref{tab:logreg_hyper} includes the final hyperparameter selection for the experimental results in §\ref{sec:lr}. We used \cite{wandb} to perform a grid search with 100 iterations, where the batch size choice was between [128, 512, 1024, 2048]. For the learning rate scheduler, we reduced the learning rate at the epoch where the NLL starts to increase. For FoMoH-3D we multiplied the learning rate by $0.8$, whereas for the other approaches we multiplied the learning rate by $0.1$. There is likely room for improvement on the parameters of the learning rate scheduler, but that would only improve the current results.

\begin{figure}[h!]
    \centering
        \centering
        \includegraphics[width=\textwidth]{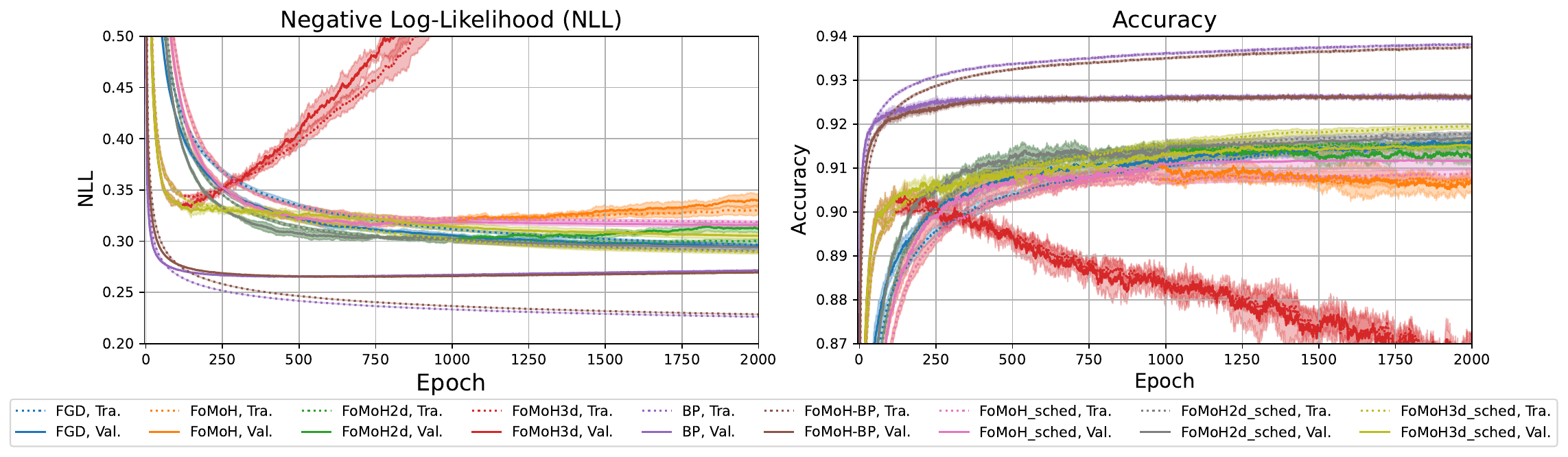}
        \caption{Training and validation curves for logistic regression model on the MNIST dataset. Average and standard deviation is shown for five random initializations.}
        \label{fig:LR}
\end{figure}

Figure \ref{fig:LR} includes the reverse-mode training and validation curves for Backpropagation and FoMoH-BP in addition to the curves shown in Figure \ref{fig:FM-LR}.

\begin{table}[!h]
\caption{Hyperparameter Optimization for Logistic Regression.}
\label{tab:logreg_hyper}
\begin{center}
\begin{scriptsize}
\begin{sc}
\begin{tabular}{lccc}
\toprule
Approach & Learning Rate & Learning Rate Bounds & Batch Size \\
\midrule
FGD  & 0.00006497 & [0.00001, 0.1] & 128 \\
FoMoH & 0.1362 & [0.001, 1.0] & 1024 \\
FoMoH (LR-Sch.)  & 0.1362 & [0.001, 1.0] & 1024 \\
FoMoH-2D  & 0.04221 & [0.001, 1.0] & 512 \\
FoMoH-2D (LR-Sch.)  & 0.04221 & [0.001, 1.0] & 512 \\
FoMoH-3D & 0.1 & [0.001, 1.0] & 512 \\
FoMoH-3D (LR-Sch.)  & 0.1 & [0.001, 1.0] & 512 \\
\midrule
FoMoH-BP & 0.04688 & [0.01, 1.0] & 2048 \\
Backpropagation  & 0.03561 & [0.01, 0.5] & 2048 \\
\bottomrule
\end{tabular}
\end{sc}
\end{scriptsize}
\end{center}
\end{table}

\subsection{CNN}\label{app:cnn}

Table \ref{tab:cnn_hyper} includes the final hyperparameter selection for the experimental results in §\ref{sec:cnn}. We used \cite{wandb} to perform Bayesian optimization with 100 iterations, where the batch size choice was fixed to 2048. For FoMoH-3D, we used the same hyperparameters as for FoMoH-2D as this gave sufficient performance (and still outperformed the other forward-mode approaches). All learning rate schedulers reduced the learning rate by 10 every 1000 epochs.

\begin{figure}[h!]
    \centering
        \centering
        \includegraphics[width=\textwidth]{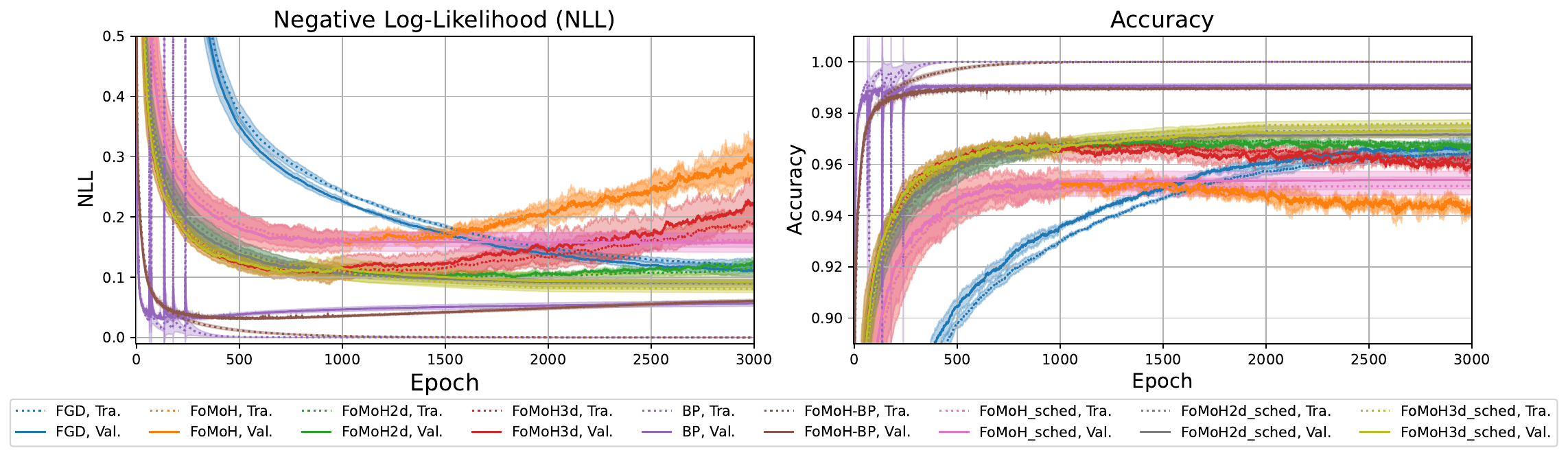}
        \caption{Training and validation curves for CNN on the MNIST dataset. Average and standard deviation is shown for three random initializations.}
        \label{fig:CNN}
\end{figure}

Figure \ref{fig:CNN} includes the reverse-mode training and validation curves for Backpropagation and FoMoH-BP in addition to the curves shown in Figure \ref{fig:FM-CNN}.

\begin{table}[!h]
\caption{Hyperparameter Optimization for Logistic Regression.}
\label{tab:cnn_hyper}
\begin{center}
\begin{scriptsize}
\begin{sc}
\begin{tabular}{lccc}
\toprule
Approach & Learning Rate & Learning Rate Bounds & Batch Size \\
\midrule
FGD  & 0.0001376 & [0.00001, 0.1] & 2048 \\
FoMoH & 0.542 & [0.001, 1.0] & 2048 \\
FoMoH (LR-Sch.)  & 0.542 & [0.001, 1.0] & 2048 \\
FoMoH-2D  & 0.3032 & [0.001, 1.0] & 2048 \\
FoMoH-2D (LR-Sch.)  & 0.3032 & [0.001, 1.0] & 2048 \\
FoMoH-3D & 0.3032 & [0.001, 1.0] & 512 \\
FoMoH-3D (LR-Sch.)  & 0.3032 & [0.001, 1.0] & 2048 \\
\midrule
FoMoH-BP & 0.04688 & [0.01, 1.0] & 2048 \\
Backpropagation  & 0.03561 & [0.005, 0.2] & 2048 \\
\bottomrule
\end{tabular}
\end{sc}
\end{scriptsize}
\end{center}
\end{table}

\subsection{Computational Resources}\label{app:resource}
All experiments are run on a NVIDIA RTX 6000 GPU. The main compute cost came from both the grid search and the Bayesian optimization that we ran over the six different optimization routines for the different experiments.